\documentclass[psfig,10pt]{asme2ej}
\usepackage{enumerate}
\usepackage{amsfonts}
\usepackage{graphicx}
\usepackage{amsfonts}
\usepackage{graphicx}
\usepackage{amsmath, amssymb, bm}
\usepackage[fixamsmath,disallowspaces]{mathtools}
\usepackage{xcolor}
\usepackage{fixmath}
\usepackage{pstool}
\usepackage{siunitx}
\usepackage{caption}
\usepackage{subcaption}

\confshortname{IDETC/CIE 2024}
\conffullname{The ASME 2024 International Design Engineering Technical Conferences \&\\
              Computers and Information in Engineering Conference}
\confdate{25-28}
\confmonth{August}
\confyear{2024}
\confcity{Washington}
\confcountry{DC}
\papernum{}

\newtheorem{definition}{Definition}

\begin{document}

\parindent0pt \setcounter{topnumber}{9} \setcounter{bottomnumber}{9} %
\renewcommand{\textfraction}{0.00001}

\renewcommand {\floatpagefraction}{0.999} \renewcommand{\textfraction}{0.01} %
\renewcommand{\topfraction}{0.999} \renewcommand{\bottomfraction}{0.99} %
\renewcommand{\floatpagefraction}{0.99} \setcounter{totalnumber}{9}

\title{\vspace{-4ex}
	Simultaneous Stiffness and Trajectory Optimization for Energy Minimization of Pick-and-Place Tasks of SEA-Actuated Parallel Kinematic Manipulators
} 
\author{
	{\underline{Thomas Kordik}, Hubert Gattringer, Andreas M\"uller}
	\affiliation{Johannes Kepler University Linz, Austria\\
		thomas.kordik@jku.at}
}


\maketitle 

\begin{abstract}
A major field of industrial robot applications deals with repetitive tasks that alternate between operating points. For these so-called pick-and-place operations, parallel kinematic manipulators (PKM) are frequently employed. These tasks tend to automatically run for a long period of time and therefore minimizing energy consumption is always of interest. Recent research addresses this topic by the use of elastic elements and particularly series elastic actuators (SEA). This paper explores the possibilities of minimizing energy consumption of SEA actuated PKM performing pick-and-place tasks. The basic idea is to excite eigenmotions that result from the actuator springs and exploit their oscillating characteristics. To this end, a prescribed cyclic pick-and-place operation is analyzed and a dynamic model of SEA driven PKM is derived. Subsequently, an energy minimizing optimal control problem is formulated where operating trajectories as well as SEA stiffnesses are optimized simultaneously. Here, optimizing the actuator stiffness does not account for variable stiffness actuators. It serves as a tool for the design and dimensioning process. The hypothesis on energy reduction is tested on two (parallel) robot applications where redundant actuation is also addressed. The results confirm the validity of this approach.
\end{abstract}

\setcounter{topnumber}{3} \setcounter{bottomnumber}{2} \renewcommand{%
\textfraction}{0.00001}

\section{INTRODUCTION}
Industrial applications often rely on robots that are able to perform fast and repetitive motions and simultaneously ensure positioning accuracy, i.e., pick-and-place-like tasks. To this end, parallel kinematic manipulators (PKM) were developed where actuation redundancy can be utilized to further enhance the robot's accuracy, speed, payload and stiffness \cite{Merlet2006, Gosselin2018}. Past research has focused mainly on time-optimal motion planning. Aiming at more resourceful and sustainable production, energy optimal planning and control is becoming increasingly important, however. Besides energy-optimal trajectory planning \cite{Lorenz2017} or exploiting actuation redundancy \cite{Lee2015}, several other strategies can be pursued. One is to replace robot components with alternative lightweight materials \cite{Briot2020}.
A more feasible concept is to introduce temporary energy storages, i.e. elastic elements and exploit the resulting inherent dynamics \cite{Barreto2016, Mirz2018}. Relevant candidates explored in the past are elastic actuators \cite{Verstraten2016}, where a spring is introduced either parallel to the motor (PEA) \cite{Haeufle2012} or in series (SEA) between the motor and load side of the actuator \cite{Pratt1995}. These actuators have been developed further so that the actuator spring stiffness can be controlled and changed during its use. This kind of actuator is also called variable stiffness actuator (VSA). The latest contributions to the topic of (variable stiffness) springs in parallel to actuation \cite{Hill2018, Hill2020, Hill2022} suggest that significant amounts of energy can be saved even for rather fast pick-and-place trajectories. To the best of the authors' knowledge, the idea of combining elastic actuators and PKM was mostly limited to applications where only PEAs were applied. Motivated by the results presented in \cite{Verstraten2016} and further advances shown in \cite{Krivosej2023}, this paper explores the possibilities of energy minimization for repetitive motion tasks where only SEAs are used as drives in fully parallel kinematic mechanisms \cite{Merlet2006}.
The paper is organized as follows. In Sec. \ref{SecII_I} a pick-and-place task is described in terms of its key characteristics, and the optimality criteria are introduced. To motivate the use of SEAs, an introductory example is discussed in Sec. 3, where different actuation schemes are compared. For this simple example it is shown that a significant energy saving is possible by means of (possibly redundant) actuation via SEAs. 
The general energy minimizing optimal control and design problem is introduced in Sec. 4 along with the dynamics model and a discussion on the differential flatness of the control problem. The numerical solution is addressed in Sec. 5, where results are shown for a 2-DOF fully parallel planar robot. It is shown, by means of this example, that significant energy reduction is possible depending on the mass and friction distribution among the PKM and the drives. Sec. 6 concludes the paper with a few suggestions for future research.

\section{Prerequisite Definitions}
\subsection{Description of Pick-and-Place Operation}\label{SecII_I}
\begin{figure}[tb]
	\centering
	\includegraphics[width=0.75\linewidth]{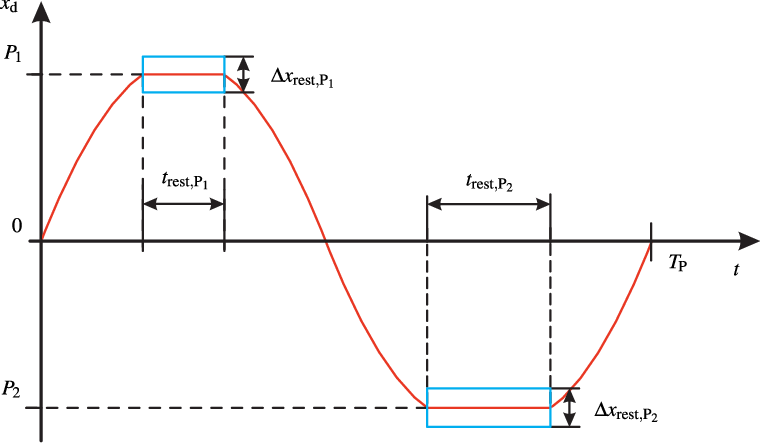}
	\caption{Desired motion of ideal pick-and-place like operation}
	\label{fig::harmonic}
\end{figure}
A typical pick-and-place task can be described as a repetitive, often periodic operation. As an example for the 1D-case, a possible desired motion $x_\mathrm{d}$ is depicted in Fig. \ref{fig::harmonic}. The task can be described by a harmonic oscillation between two operation points $P_1$ and $P_2$ where the robot's tool-center-point rests for a short time. These resting phases, in which a workpiece is gripped or released, e.g., sorting on a conveyor belt, are denoted with $t_{\mathrm{rest,P_1}}$ and $t_{\mathrm{rest,P_2}}$, respectively. Since modern grippers are able to perform high-speed grip and release motions, it is sufficient that the tool-center-point does not come to rest but stays within certain allowed displacement windows denoted with $\Delta x_\mathrm{rest,P_1}$ and $\Delta x_\mathrm{rest,P_2}$. In general, different requirements for the resting times $t_\mathrm{rest,P_1}$ and $t_\mathrm{rest,P_2}$ can be demanded as gripping and releasing motions itself may take diverse time spans. In conclusion, a cyclic pick-and-place operation can be fully defined by its cyclic time $T_\mathrm{P}$, its operating points $P_i$ as well as its respective resting windows, depicted as blue rectangles in Fig. \ref{fig::harmonic}, consisting of resting times $t_\mathrm{rest,P_i}$ and allowed deviations $\Delta x_\mathrm{rest,P_i},\ \  i = 1,2$. The trajectory connecting these resting windows can take any shape as long as periodicity is ensured. 

%
%
%
\subsection{Optimality Criteria}
In general, the goal of energy efficient or energy optimal trajectory planning is to minimize electrical energy consumed by actuators, i.e., motors, of a robot system. This heavily depends on the used motor type, control unit and availability of energy buffers and recuperation features. Besides mechanical output energy of an actuator
\begin{equation}
	E_{\mathrm{output},i} = \int P_{\mathrm{output},i}\,\mathrm{d}t = \int \tau_i\, \dot{q}_i\, \mathrm{d}t
\end{equation}
with mechanical output power $P_{\mathrm{output},i}$, actuation torque/force $\tau_i$ and velocity $\dot{q}_i$, sources of energy loss are resistive, conduction and switching losses \cite{Hill2020}. As the last two mentioned can mostly be omitted, the electrical energy loss per drive is defined as
\begin{equation}
E_{\mathrm{el,loss},i} = \int P_{\mathrm{el,loss},i}\, \mathrm{d}t = \int \frac{\tau_i^2}{k^2_{m,i}} R \mathrm{d}t
\end{equation}
with the electrical power loss $P_{\mathrm{el,loss},i}$, the resistance of the stator winding $R$ and assumed linear motor torque constant $k_{m,i}$. This results in the model for the consumed electrical power of one drive
\begin{equation}
	P_{\mathrm{el},i} =  P_{\mathrm{output},i} + P_{\mathrm{el,loss},i}.
\end{equation}
Recuperation into the grid or temporary energy buffers requires advanced elctronic devices. Nevertheless, motor control units operating $n$ motors at once can recuperate energy by small margins. Here, energy generated during braking of one motor can be reused to actuate another. Surplus energy is otherwise dissipated. The energy optimal cost criteria \cite{Mirz2018} is subsequently given by
\begin{equation}
	\min E_{\mathrm{el}} = \min \int P_{\mathrm{el}}^+\, \mathrm{d}t,\label{equ::Eloss}
\end{equation}
with
\begin{equation}
	P_{\mathrm{el}}^+ = \left\{\begin{array}{rr}
		\sum_{i=1}^{n} P_{\mathrm{el},i} &\ \text{for}\ \ \sum_{i=1}^{n} P_{\mathrm{el},i} > 0\\
		 0 &\ \text{for}\ \ P_{\mathrm{el},i} \leq 0
		\end{array}\right.. \label{equ::CostFcn}
\end{equation}
\section{Motivating Example}\label{sec::SecIII}
\begin{figure}[b]
	\centering
	\includegraphics[width=0.84\linewidth]{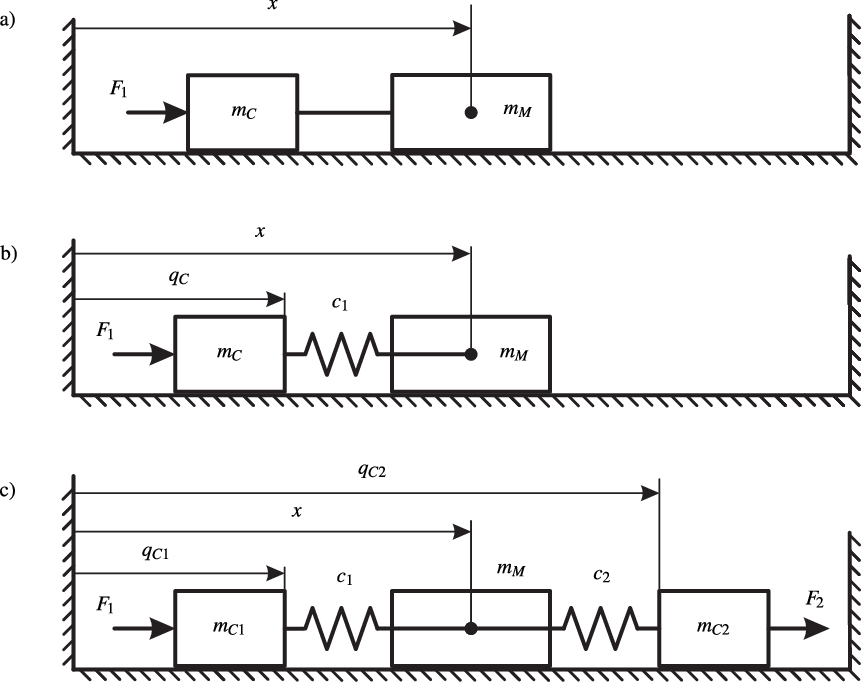}
	\caption{Considered setups in the introductory example: a) Setup I - Fully actuated mechanism with rigid drives (FA-RD), b) Setup II - Fully actuated mechanism with SEA (FA-SEA), c) Setup III - Redundantly actuated mechanism with SEA (RA-SEA)}
	\label{fig::LinearAxis}
\end{figure}

We consider the simple example of a mechanism where an actuated slider with mass $m_M$ moves on a horizontal linear guide. The slider is to be moved periodically with a given frequency between two operating points as described in Sec. \ref{SecII_I}. Three setups with different means of actuation, shown in Fig. \ref{fig::LinearAxis}, are considered:
\begin{enumerate}[I]
	\item One actuator rigidly attached to the slider
	\item One actuator connected to the slider by a spring
	\item Two actuators connected to the slider by springs
\end{enumerate}
We assume that an energy optimal control problem can be formulated, finding the optimal connecting trajectory and, if available, the optimal elastic actuator elements  to minimize energy consumption according to equation (\ref{equ::Eloss}). This is then solved for all three setups and the results are compared.

\subsection{Dynamics for Different Actuation Schemes}
\subsubsection{Fully Actuated Mechanism with Rigid Drives}
The actuation scheme, shown in Fig. \ref{fig::LinearAxis}a, is discussed first.
The slider is fixed rigidly to an actuator with mass $m_C$. Its position is described by $x$ and is fully actuated by the input force $F$. This system will be called \textit{fully actuated mechanism with rigid drives (FA-RD)}. Its equation of motion is
\begin{equation}
(m_M+m_{C}) \ddot{x} + d_v \dot{x} = F,
\end{equation}
where $d_v$ is the viscous friction coefficient.

\subsubsection{Fully Actuated Mechanism with SEA}
Replacing the rigid drive in setup I by a SEA with a constant stiffness $c$ yields setup II, which is shown in Fig. \ref{fig::LinearAxis}b. The mass $m_M$ is again fully actuated by the SEA. However, the consequence of this type of actuation is that the combined system becomes underactuated and is governed by 
\begin{equation}
	\begin{bmatrix}m_M&0\\0&m_{C}\end{bmatrix} \begin{bmatrix}\ddot{x}\\\ddot{q}_C\end{bmatrix}+\begin{bmatrix}c (x-q_C)\\-c (x-q_C) + d_v \dot{q}_C\end{bmatrix} = \begin{bmatrix}0\\F\end{bmatrix}.
\end{equation}
Only the drive $m_C$ is assumed to be subject to viscous friction $d_v$. Minimal coordinates are given by $x$ and $q_C$ and the only input force by $F$. As the underactuation solely results from the SEA dynamics while the mass $m_M$ itself is actuated by as many actuators as it features degrees of freedom, we call this setup due to the parallels to the previous section \textit{fully actuated mechanism with SEA (FA-SEA)}.
\subsubsection{Redundantly Actuated Mechanism with SEA}
Adding another linear SEA drive leads to setup III, depicted in Fig. \ref{fig::LinearAxis}c. The system now exhibits a parallel topology and is for the same reason as setup II underactuated. While for rigid actuators mass $m_M$ would have been redundantly actuated and because underactuation again solely results from the type of actuation, setup III is called \textit{redundantly actuated mechanism with SEA (RA-SEA)}.
The drive masses are denoted with $m_{C1}$ and $m_{C2}$ and their position is described by $q_{C1}$ and $q_{C2}$ respectively. Both are connected to the mass $m_M$ by linear springs with constant stiffnesses $c_1, c_2$ and the mass position is again given by $x$. The dynamic equations of motion are 
\begin{equation}
		\begin{bmatrix}m_M&0&0\\0&m_{C1}&0\\0&0&m_{C2}\end{bmatrix}\begin{bmatrix}\ddot{x}\\\ddot{q}_{C1}\\\ddot{q}_{C2}\end{bmatrix} + \\ \left[\begin{array}{c}
	c_1 \left(x - q_{C1}\right) + c_2 \left(x - q_{C2}\right) 
	\\
	- c_1 \left(x - q_{C1}\right) + d_{v} \dot{q}_{C1}\\
	- c_2 \left(x - q_{C2}\right) + d_{v} \dot{q}_{C2} 
	\end{array}\right]= \begin{bmatrix}0\\F_1\\F_2\end{bmatrix},
\end{equation}
where $F_1$ and $F_2$ represent the actuation forces.
\subsection{Optimal Results for Exemplary Task}
\begin{figure}[h]
	\centering	
	\includegraphics{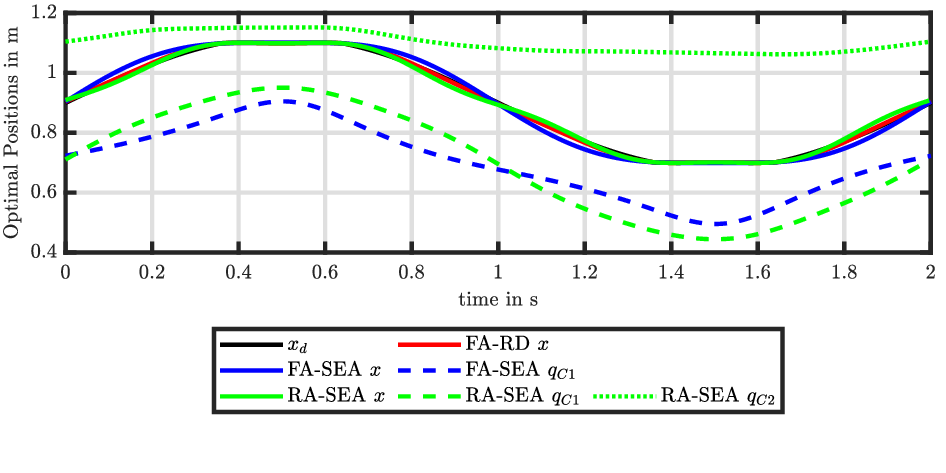}
	\vspace{-0.7cm}
	\caption{Linear Introductory Example: Comparison of optimal slider and drive trajectories}
	\label{fig::LinearAxisPositions}
\end{figure}

\begin{figure}[htb]
	\centering	
	\includegraphics{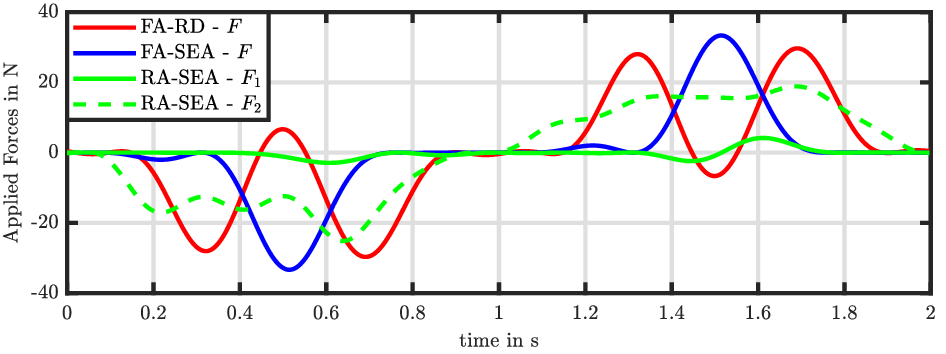}
	\caption{Linear Introductory Example: Comparison of optimal actuation forces}
	\label{fig::LinearAxisForces}
\end{figure}

\begin{figure}[htb]
	\centering	
	\includegraphics{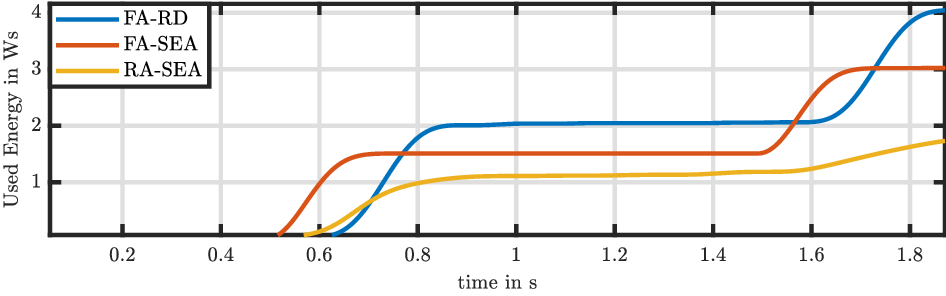}
	\caption{Linear Introductory Example: Minimized energy consumed during one cycle of a pick-and-place operation}
	\label{fig::LinearAxisEnergy}
\end{figure}

To showcase the possibility of saving energy in pick-and-place operations, all three setups are subject to an exemplary task. It is presumed that an optimization problem for two operating points $P_1 = \SI{0.7}{\meter}$ and $P_2 = \SI{1.1}{\meter}$, defining a cyclic motion with cycle time $T_\mathrm{P} = \SI{2}{\second}$, can be formulated. Symmetrical resting times $t_{\mathrm{rest,P_1}} = t_{\mathrm{rest,P_2}} = \SI{275}{\milli \second}$ as well as allowed deviations $\Delta x_{\mathrm{rest,P_1}} = \Delta x_{\mathrm{rest,P_2}} = \SI{1}{\milli \meter}$ are demanded. For dynamic simulation parameters $m_M = \SI{2}{\kilogram}, m_C = m_{C1} = m_{C2} = \SI{5}{\kilogram}, d_v = \SI{2}{\newton \meter/\second}$ were identified for a linear axis experimental prototype and physically meaningful dynamic limits are chosen.  Because cyclic motions are observed and oscillations due to springs are expected (see Sec. \ref{sec::SecV} for details), the trajectory for mass $m_M$ is assumed to be harmonic and expressed using a Fourier series
\begin{equation}
x(t) = x_0 + \sum_{l=1}^{10}(A_l \sin(l\omega t) + B_l \cos(l\omega t)). \label{equ::Fourier}
\end{equation}
Therefore, Fourier coefficients $A_l, B_l, l = 1..10$, an offset $x_0$, and if present, actuator spring stiffnesses, serve as optimization variables in the process. A detailed description of the implemented optimization problem is stated in Sec. \ref{sec::SecIV_III}. Figures \ref{fig::LinearAxisPositions}-\ref{fig::LinearAxisEnergy} show the optimal solutions for each setup according to the cost function (\ref{equ::Eloss}). Besides the desired mass trajectory $x_d$, the optimized positions of $m_M$ and the actuators (dashed lines) are shown in Fig. \ref{fig::LinearAxisPositions}. The required forces are shown in Fig. \ref{fig::LinearAxisForces} and the consumed electrical energy over one period is displayed in Fig. \ref{fig::LinearAxisEnergy}. A numerical comparison in terms of absolute and relative values of consumed energy with respect to FA-RD is presented in Table \ref{tab::LinearAxisEnergy}. For this simple example, it becomes clear that by introducing a serial spring with an optimal spring stiffness which matches the desired trajectory frequencies, significant amounts of energy ($>25\%$) can be saved. This study also indicates that with redundant actuation the consumed energy can further be reduced drastically as eigenfrequencies can be better matched. Table \ref{tab::LinearAxisEnergy} additionally displays the optimal spring stiffnesses $c_i$ of the SEA. In case the resting times were set to zero and friction was omitted, the optimal spring stiffnesses would be the ones that match the system's eigenfrequency with the task frequency $1/T_\mathrm{P}$. Then, natural oscillations could perform the desired task and energy consumption would shrink to zero. So, in general, for resting times $t_{\mathrm{rest,P_i}} > \SI{0}{\second}$, optimal $c^\ast_i$ rather take those values that best match relevant occurring frequencies .

\begin{table}[htb]
		\caption{Linear Introductory Example: Minimal energy consumption and optimal SEA stiffnesses for a pick-and-place operation. The relative values are normalized with respect to FA-RD.}
		\vspace{-2ex}
	\begin{center}
		\begin{tabular}{|c||c|c|c|}
			\hline
			&$E_{\mathrm{el}}$ in $\SI{}{\watt\second}$&$E_{\mathrm{el}}/E_{\mathrm{el,FAMRD}}$ &$c^\ast_i$ in $\SI{}{\newton \meter/\radian}$\\
			\hline
			FA-RD&$4.07$&/&/\\
			\hline
			FA-SEA&$3.02$& $74.2\,\%$&$c^\ast = 76.5$\\
			\hline
			RA-SEA&$1.79$& $44.0\,\%$&\begin{tabular}{@{}c@{}}$c^\ast_1 = 307$\\ $c^\ast_2 = 94.5$\end{tabular}\\
			\hline
		\end{tabular}
	\end{center}
	\label{tab::LinearAxisEnergy}
	\vspace{-2ex}
\end{table}

\section{Energy Minimizing Optimal Control}\label{sec::SecIV}
\begin{figure}[h]
	\centering
	\includegraphics[width=0.95\textwidth]{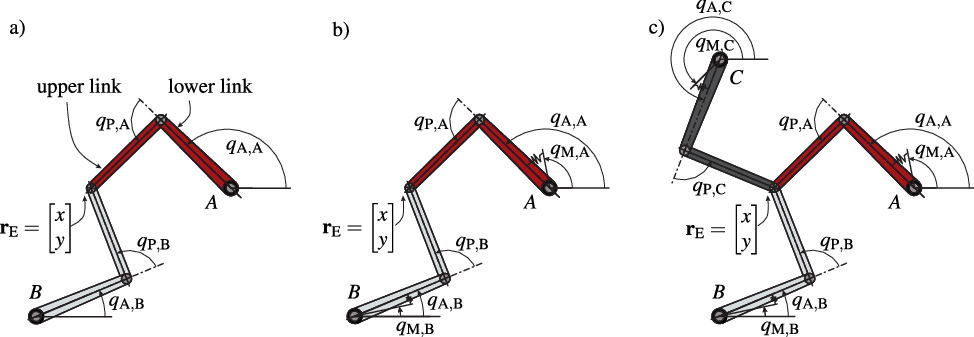}
		\caption{Observed PKM setups analogous to Fig. 2: a) Setup I - PKM RD, b) Setup II - PKM SEA, c) Setup III - RAPKM SEA}
		\label{fig::PKM3ArmSEA}
\end{figure}

Motivated by the results of the previous section, a 2 DOF planar parallel robot with three identical arms enumerated by $i = A,B,C$, shown in Fig. \ref{fig::PKM3ArmSEA}c, is analyzed and possibilities of energy savings during pick-and-place tasks are examined. Here, a three-armed fully parallel planar mechanism actuated by stationary mounted SEAs is shown. Analogously to the motivating example, three setups for comparison, depicted in Fig. \ref{fig::PKM3ArmSEA}, can be derived. For better understanding, some prerequisites need to be addressed first. Well-established classifications for PKM that are actuated by rigid drives (no compliance of gear or motor unit) and composed of only rigid bodies can be found in \cite{Mueller2013, Muller2005}. Actuation is classified as follows.

\begin{definition}
	Consider a parallel mechanism comprising rigid bodies only. Denote with $\delta $ the DOF of the mechanism without drives, and with $m$ the number of actuators. The mechanism is \emph{redundantly actuated} iff $\delta < m$ (more independent inputs than the mechansim DOF), \emph{fully actuated} iff $\delta = m$, and \emph{underactuated} iff $\delta > m$.
\end{definition}
Notice that in this definition the mechanism does not include the drives. Thus, when it is referred to as either PKM or redundantly actuated PKM, the mechanism itself (use of two/three arms) is concerned. Completely replacing all rigid drives by SEAs results in an underactuated system. We distinguish between the following three situations:
\begin{enumerate}[I]
	\item Non-redundantly actuated PKM (arms $A$ and $B$ used) with rigid drives (\textit{PKM RD})
	\item Non-redundantly rigid-actuated PKM (arms $A$ and $B$ used) with SEA (\textit{PKM SEA})
	\item Redundantly rigid-actuated PKM (arms $A,B$, and $C$ used) with SEA (\textit{RAPKM SEA}).
\end{enumerate}
First the dynamic models are derived and subsequently an energy minimizing optimization problem is formulated.

\subsection{Modelling of Actuated PKM}
\subsubsection{Dynamic Model of PKM Without Actuator Dynamics}\label{sec::SecIV_I_I}
An efficient approach to express the dynamics equations of motion (EOM) of a fully parallel robot is the modular modeling proposed in \cite{Mueller2020, Mueller2022} which is an extension of those presented in \cite{Angeles1988, Angeles2007, Briot2015}. The EOM of a fully parallel robot with $m$ limbs, where only the first link per limb is actuated, can be expressed by
\begin{equation}
\mathbf{M}(\mathbf{q}) \ddot{\mathbf{q}}_{\mathrm{A,ind}} + \mathbf{g}(\mathbf{q},\dot{\mathbf{q}}) = \mathbf{H}_\mathrm{A}^T(\mathbf{q}) \mathbf{u} \label{equ::EOMPKM}.
\end{equation}
With $\mathbf{q} = [\mathbf{q}_\mathrm{A}^T,\mathbf{q}_\mathrm{P}^T]$ we denote the vector of all joint coordinates which consists of actuated $\mathbf{q}_\mathrm{A}\in\mathbb{R}^m$ and passive $\mathbf{q}_\mathrm{P}$ joint coordinates. Furthermore, $\mathbf{q}_\mathrm{A,ind}\in\mathbb{R}^\delta$, a subvector of $\mathbf{q}_\mathrm{A}$, describes an independent set of actuated joint coordinates, where $\delta$ is the mechanism's DOF. The generalized mass matrix is denoted with $\mathbf{M}(\mathbf{q})\in \mathbb{R}^{\delta,\delta}$, the actuation torques/forces with $\mathbf{u}\in\mathbb{R}^m$ and all remaining forces/ torques are summarized in $\mathbf{g}(\mathbf{q},\dot{\mathbf{q}})$. As the dominant amount of friction in a robotic systems tends to occur in its actuators, friction is not included in the PKM modeling. The matrix $\mathbf{H}_\mathrm{A}(\mathbf{q})\in \mathbb{R}^{m,\delta}$ defines the kinematic relation 
\begin{equation}
\dot{\mathbf{q}}_\mathrm{A} = \mathbf{H}_\mathrm{A}(\mathbf{q})\dot{\mathbf{q}}_\mathrm{A,ind}
\end{equation}
and maps $\mathbf{u}$ onto the generalized coordinates. If a fully actuated PKM is considered, $\mathbf{H}_\mathrm{A}(\mathbf{q})$ poses as an identity matrix and all actuated joint coordinates $\mathbf{q}_\mathrm{A}$ are kinematically independent, so $\mathbf{q}_\mathrm{A,ind} = \mathbf{q}_\mathrm{A}$ holds. Otherwise, for a redundantly actuated PKM $\mathbf{q}_\mathrm{A} =\left[\mathbf{q}_\mathrm{A,ind}^T,\mathbf{q}_\mathrm{A,dep}^T\right]^T$ can be split into an independent and dependent set of actuated coordinates $\mathbf{q}_\mathrm{A,dep}$ and $\mathbf{H}_\mathrm{A}(\mathbf{q})$ is non-square.
\subsubsection{Dynamic Model of PKM Including Actuator Dynamics}
The most common approach to define the dynamics of SEAs, proposed
in \cite{Spong1987}, is to model the elastic actuator by a motor inertia and a spring
that interconnects it to its load. This model is in theory further applicable to nonlinear and variable stiffness actuators (VSA) \cite{Palli2008} but not regarded in this paper. The additional DOFs, i.e., the vector of actuator coordinates (coordinates of motor inertia) of all SEAs, are summarized in $\mathbf{q}_\mathrm{M}\in\mathbb{R}^m$. The extended vector of all joint coordinates is denoted with
$\overline{\mathbf{q}}^T = [\mathbf{q}^T_\mathrm{M},\mathbf{q}^T]$. Assuming linear spring characteristics, the actuation torques/forces in equation (\ref{equ::EOMPKM}) are expressed as
\begin{equation}
\mathbf{u} = \mathbf{C}(\mathbf{q}_\mathrm{M} - \mathbf{q}_\mathrm{A})
\end{equation}
where $\mathbf{C} = \mathrm{diag}(c_1,\ldots c_m)$, describes the compliance of the SEAs, with $c_i,\  i = 1\ldots m$ being the stiffness coefficient associated to drive $i$. Analogously, the inertia matrix of the actuators (motors) is denoted with $\mathbf{M}_\mathrm{M} = \mathrm{diag}(m_1, . . .m_m)$ and with $\bm{\tau}\in \mathbb{R}^m$ the motor torques(forces) of the drives within the SEA. Here, $m_i,\ i = 1\ldots m$ is the inertia moment of the $i$-th drive (representing motor and gear unit), which is coupled to the $i$-th actuator coordinate in $\mathbf{q}_\mathrm{A}$ via the spring with stiffness $c_i,\ i = 1\ldots m$. Viscous friction $\mathbf{d}_v = \mathrm{diag}([d_{v,1},\hdots,d_{v,m}])$ is assumed to affect only the actuator coordinates $\dot{\mathbf{q}}_\mathrm{M}$. Coulomb friction and damping in parallel to the springs is neglected. Combined with equation (\ref{equ::EOMPKM}), the EOM of the combined system - fully parallel robots actuated with SEAs - can be governed by
\begin{equation}
\begin{bmatrix} {\mathbf{M}}_\mathrm{M} & 0 \\
0  & {\mathbf{M}}(\mathbf{q})  \end{bmatrix} \begin{bmatrix} \ddot{\mathbf{q}}_\mathrm{M} \\ \ddot{\mathbf{q}}_\mathrm{A,ind}  \end{bmatrix} + \begin{bmatrix} \mathbf{d}_v \dot{\mathbf{q}}_\mathrm{M} \\ {\mathbf{g}}(\mathbf{q},\dot{\mathbf{z}})
\end{bmatrix}+ \begin{bmatrix} \mathbf{C} & -\mathbf{C}\\
-\mathbf{H}_\mathrm{A}^T(\mathbf{q})\mathbf{C}  &\mathbf{H}_\mathrm{A}^T(\mathbf{q})\mathbf{C} \end{bmatrix} \begin{bmatrix} \mathbf{q}_\mathrm{M} \\ \mathbf{q}_\mathrm{A}  \end{bmatrix} = \begin{bmatrix} \bm{\tau} \\ \mathbf{0}  \end{bmatrix}.\label{equ::EOMAll}
\end{equation}
The dynamic decoupling of the EOM for the parallel mechanism and elastic actuation is clearly visible. There is no internal dynamics and both components are solely coupled by the SEA spring stiffnesses.\\
When using standard rigid drives, the actuation torques(forces) become $\mathbf{u} = \bm{\tau}$, friction affects $\mathbf{q}_\mathrm{A}$ and the motor coordinates $\mathbf{q}_\mathrm{M}$ are dropped yielding
\begin{equation}
	(\mathbf{M}_\mathrm{M} + \mathbf{M}(\mathbf{q})) \ddot{\mathbf{q}}_{\mathrm{A,ind}} +\mathbf{H}_\mathrm{A}^T(\mathbf{q}) \mathbf{d}_v \dot{\mathbf{q}}_\mathrm{A} + \mathbf{g}(\mathbf{q},\dot{\mathbf{q}}) = \mathbf{H}_\mathrm{A}^T(\mathbf{q}) \bm{\tau}. \label{equ::EOMRD}
\end{equation}
\subsection{Differential Flatness}
A property that systems governed by equation (\ref{equ::EOMAll}) exhibit is differential flatness. A system is per definition differentially flat if it meets the following conditions \cite{Levine2010}.
\begin{definition}
	Assume a system with the state $\mathbf{x} \in \mathbb{R}^n$ and the input $\mathbf{u} \in \mathbb{R}^m$. This system is differentially flat if there exists a so-called flat output $\mathbf{y} \in \mathbb{R}^m$ which is locally independent and both, the system state $\mathbf{x}$ and the system input $\mathbf{u}$, can be expressed as functions of the flat output $\mathbf{y}$ and its time derivatives $\dot{\mathbf{y}},\ddot{\mathbf{y}},\dddot{\mathbf{y}}, \hdots$.
\end{definition}
A valid choice for the flat output of the fully actuated PKM with SEAs of Fig. \ref{fig::PKM3ArmSEA}b is its end-effector position $\mathbf{y} = \mathbf{r}_\mathrm{E}$. It exhibits a vector relative degree \cite{Isidori1995} of $\{4,4\}$, which indicates the number of time derivatives of $y_i$ that have to be calculated until a system input occurs first. In contrast, the redundantly actuated PKM with SEAs of Fig. \ref{fig::PKM3ArmSEA}c exhibits a vector relative degree of $\{4,4,2\}$. Thus, to completely describe the system, an additional flat output must be defined. One way to address this is to add $m-\delta = 1$ coordinates of $\mathbf{q}_\mathrm{M}$ to the flat output. An obvious choice is to use the motor coordinate of arm $A$, yielding $\mathbf{y}^T = [\mathbf{r}^T_\mathrm{E}, q_\mathrm{M,A}]$.
The property of differential flatness is rather beneficial. This allows to represent the system dynamics (\ref{equ::EOMAll}) by chains of integrators which is used in the following energy minimizing optimal control problem. Here, all PKM joint coordinates $\mathbf{q}$ can be expressed by the flat output $\mathbf{y}$ using analytical inverse kinematics and furthermore $\dot{\mathbf{q}}$ as well as $\ddot{\mathbf{q}}$ can be calculated by derivation with respect to time. Subsequently, the lower equation of (\ref{equ::EOMAll}) is solved for the motor angles yielding an expression $\mathbf{q}_\mathrm{M} = \mathbf{f}_\mathrm{M}(\mathbf{y}, \dot{\mathbf{y}}, \ddot{\mathbf{y}})$. Deriving this expression twice with respect to time and inserting it in the upper equation of (\ref{equ::EOMAll}) further allows to describe the input torques by only the flat output and its time derivatives as $\bm{\tau} = \mathbf{f}_{\bm{\tau}}(\mathbf{y}, \dot{\mathbf{y}}, \ddot{\mathbf{y}},\dddot{\mathbf{y}},\mathbf{y}^{(4)})$. This procedure also shows that damping in parallel to the SEA springs needs to be omitted as otherwise the system (\ref{equ::EOMAll}) is not anymore differentially flat.

\subsection{Energy Minimizing Optimal Control Problem}\label{sec::SecIV_III}
In this section, an energy minimizing optimal control problem is formulated for cyclic pick-and-place operations of the redundantly actuated PKM with SEAs shown in Fig. \ref{fig::PKM3ArmSEA}. From there, the optimal control problem for the fully actuated PKM with SEAs can easily be derived. A 2D-end-effector trajectory $\mathbf{r}_\mathrm{d,E} = [x_\mathrm{d,E}, y_\mathrm{d,E}]^T$ is defined where both of its coordinates conform with the description provided in Sec. \ref{SecII_I}. The strategy is to simultaneously optimize the robots' end-effector trajectory $\mathbf{r}_\mathrm{E}$ as well as the vector of SEA spring stiffnesses $\mathbf{c} = [c_1,\ldots, c_3]^T$ to hopefully exploit the SEAs' harmonic frequencies and the natural dynamics for energy saving. Thus, an assumption on the shape of the real end-effector position $\mathbf{r}_\mathrm{E}$ has to be made first. For setups with rather linear kinematics and dynamics, $\mathbf{r}_\mathrm{E}$ can be defined using Fourier series as in Sec. \ref{sec::SecIII} or polynomials, as discussed in \cite{Hill2018}. As for this PKM, the workspace exhibits highly nonlinear and state dependent characteristics, the end-effector position is assumed to be piecewise constant in the fifth time derivative of $\mathbf{r}_\mathrm{E}$, $\mathbf{r}^{(5)}_\mathrm{E}$. Here, the benefit of differential flatness becomes visible. Using coordinate transformations \cite{Levine2010}, we can represent the system dynamics (\ref{equ::EOMAll}), as shown in the previous section, by chains of integrators. Since its vector relative degree $\{4,4,2\}$ is not homogenous, two separate integrator chains for the flat outputs, $\dot{\mathbf{x}}_{\mathbf{r}_\mathrm{E}} = \mathbf{f}_{\mathbf{r}_\mathrm{E}}(\mathbf{x}_{\mathbf{r}_\mathrm{E}}, \mathbf{v}_{\mathbf{r}_\mathrm{E}})$ with the integrator state vector $\mathbf{x}_{\mathbf{r}_\mathrm{E}} = [\mathbf{r}^T_\mathrm{E},\dot{\mathbf{r}}^T_\mathrm{E},\ddot{\mathbf{r}}^T_\mathrm{E},\mathbf{r}^{(3),T}_\mathrm{E},\mathbf{r}^{(4),T}_\mathrm{E}]^T$ and input $\mathbf{v}_{\mathbf{r}_\mathrm{E}} = \mathbf{r}^{(5)}_\mathrm{E}$ as well as $\dot{\mathbf{x}}_{q_\mathrm{M,A}} = \mathbf{f}_{q_\mathrm{M,A}}(\mathbf{x}_{q_\mathrm{M,A}}, v_{q_\mathrm{M,A}})$ with $\mathbf{x}_{\mathbf{q}_\mathrm{M,A}} = \left[{q}_\mathrm{M,A},\dot{{q}}_\mathrm{M,A},\ddot{{q}}_\mathrm{M,A}\right]^T$ and $v_{q_\mathrm{M,A}} = q_\mathrm{M,A}^{(3)}$, are defined. The pick-and-place operation is only observed for one period $T_\mathrm{P}$ and the cost function used is given by equation (\ref{equ::CostFcn}).
The energy minimizing optimal control problem for the redundantly actuated PKM with SEAs can consequently be stated as
\begin{alignat}{3}
&\min_{\mathbf{x}_{\mathbf{r}_\mathrm{E}}, \mathbf{v}_{\mathbf{r}_\mathrm{E}},\mathbf{x}_{q_\mathrm{M,A}}, v_{q_\mathrm{M,A}},\mathbf{c}}  \int_{t=0}^{T_\mathrm{P}} P_\mathrm{el}\,\mathrm{d}t\span\span\span\span\\
&\mathrm{s.t.} \quad && \dot{\mathbf{x}}_{\mathbf{r}_\mathrm{E}} = \mathbf{f}_{\mathbf{r}_\mathrm{E}}(\mathbf{x}_{\mathbf{r}_\mathrm{E}}, \mathbf{v}_{\mathbf{r}_\mathrm{E}})\label{equ::DynBegin} \\
&&& \dot{\mathbf{x}}_{q_\mathrm{M,A}} = \mathbf{f}_{q_\mathrm{M,A}}(\mathbf{x}_{q_\mathrm{M,A}}, v_{q_\mathrm{M,A}})\label{equ::DynMiddle}\\
&&& \bm{\tau} = \mathbf{f}_{\bm{\tau}}(\mathbf{c},\mathbf{x}_{\mathbf{r}_\mathrm{E}},\mathbf{x}_{q_\mathrm{M,A}}) \label{equ::DynEnd}\\
&&& |\mathbf{r}_{\mathrm{E,d}} - \mathbf{r}_\mathrm{E}|(t_{\mathrm{rest,P_j}}) \leq \Delta \mathbf{r}_{\mathrm{rest,P_j}}, \quad  j = 1,2\label{equ::DesPath}\\
&&& \mathbf{x}_{\mathbf{r}_\mathrm{E}}(0) = \mathbf{x}_{\mathbf{r}_\mathrm{E}}(T_\mathrm{P})\label{equ::Period}\\
&&& \mathbf{x}_{q_\mathrm{M,A}}(0) = \mathbf{x}_{q_\mathrm{M,A}}(T_\mathrm{P})\label{equ::Period2}\\
&&& 0 < c_i, \quad i = 1 \ldots 3.\label{equ::OptEnd}
\end{alignat}
Here, equations (\ref{equ::DynBegin}) - (\ref{equ::DynMiddle}) account for the integrator chains and equation (\ref{equ::DynEnd}) for the inverted EOM (\ref{equ::EOMAll}) where the motor torques $\bm{\tau}$ can be expressed as a function of the flat output and its time derivatives. The characterizing constraint (\ref{equ::DesPath}) ensures that the end-effector position $\mathbf{r}_\mathrm{E}$ stays within the boundaries of the prescribed pick-and-place resting windows, shown in Fig. \ref{fig::harmonic}. To guarantee periodicity, equations (\ref{equ::Period})  and (\ref{equ::Period2}) are included. Additionally, box constraints for the position, velocity, acceleration, etc., for both, joint and end-effector coordinates as well as motor torque constraints can be defined. In case only the fully actuated PKM, both with rigid actuators and SEAs, should be subject of the optimization, conditions (\ref{equ::DynMiddle}) and (\ref{equ::Period2}) vanish and $m = 2$. 
To numerically solve the continuous optimization problem a multiple-shooting approach with $200$ samples is used. 
Numerical integration in (\ref{equ::DynBegin}) and (\ref{equ::DynMiddle}) is done with the explicit Runge-Kutta method of fourth order and for each sample  constant integrator inputs $\mathbf{v}_{\mathbf{r}_\mathrm{E}}$ and ${v}_{q_\mathrm{M,A}}$\ are assumed. The whole optimal control problem is implemented in the CasADi framework \cite{Andersson2019} and IPOPT \cite{Waechter2006} is applied as a solver.

\section{Simulation Results of Pick-and-Place Task}\label{sec::SecV}
The energy minimizing optimal control problem is evaluated for a cyclic pick-and-place task between the operating points $\mathbf{P}_1 = [-0.07, -0.07]\,\mathrm{m}$ and $\mathbf{P}_2 = [0.07, 0.07]\,\mathrm{m}$. The task space, indicated by the black rounded triangular border in the following end-effector trajectory plots, is exploited as best as singularities of the fully actuated PKM allow for. The cycle time is chosen $T_\mathrm{P} = \SI{1}{\second}$, close to time-optimal results in \cite{Kordik2023} while still leaving enough freedom for energy minimization. The geometric and dynamic properties are taken from CAD data where the arm links are designed to be 3D-printed with polylactide (PLA). All arm links are $20\,\mathrm{cm}$ long and their masses are $110.4\,\mathrm{g}$ for the upper links and $44.8\,\mathrm{g}$ for the lower link, respectively. The used motors exhibit moments of inertia of $4.4 \times 10^{-6}\,\mathrm{kg\,m^2}$ and a maximum torque of $0.3\,\mathrm{Nm}$.
Symmetrical resting phase times $t_\mathrm{rest,P_1}=t_\mathrm{rest,P_2} = \SI{100}{\milli \second}$ as well as symmetrical deviation windows $\Delta \mathbf{r}_\mathrm{rest,P_1} = \Delta \mathbf{r}_\mathrm{rest,P_2} = \SI{1}{\milli \meter}$ are demanded. Three different studies in simulations are carried out and discussed in the following.
\subsection{Negligible Friction}\label{sec::SecV_I}
\begin{figure}[h]
	\centering	
	\includegraphics{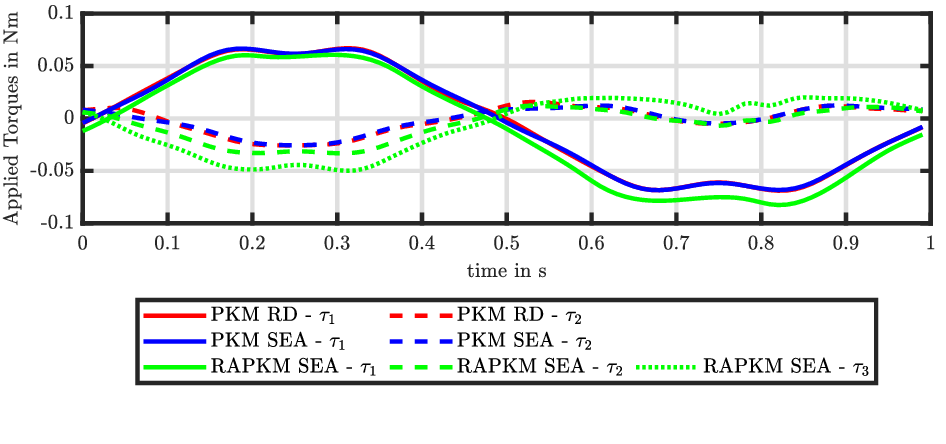}
	\vspace{-1cm}
	\caption{Energy minimized motor torques $\bm{\tau}$ of PKMs in absence of friction}
	\label{fig::PaPMLowFric}
\end{figure}

\begin{figure}[h]
	\centering	
	\includegraphics{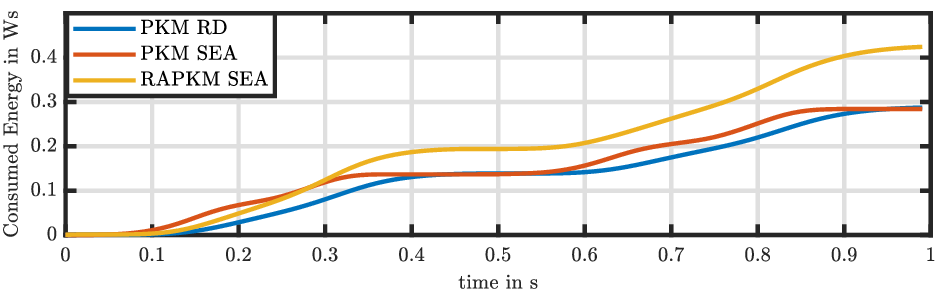}
	\caption{Consumed energy $E_\mathrm{el}$ over one cycle of PKMs in absence of friction}
	\label{fig::PaPEnergyLowFric}
\end{figure}

To observe the possible excitation of eigenfrequencies and natural motions, at first friction is assumed to be negligibly small and therefore effectively set to zero. Results are shown in Figs. \ref{fig::PaPEELowFric} - \ref{fig::PaPEnergyLowFric} and numerical values are displayed in Table \ref{tab::PaPEnergyLowFric}. Here, Fig. \ref{fig::PaPEELowFric} shows the energy minimizing trajectories for the three setups of Sec. \ref{sec::SecIV}. Figure \ref{fig::PaPMLowFric} depicts the minimized motor torques and Fig. \ref{fig::PaPEnergyLowFric} displays the consumed energy throughout the pick-and-place operation. \vspace{0.35cm}\\ 
\begin{minipage}[h]{0.39\textwidth}
	\centering	
	\includegraphics[width=0.95\textwidth]{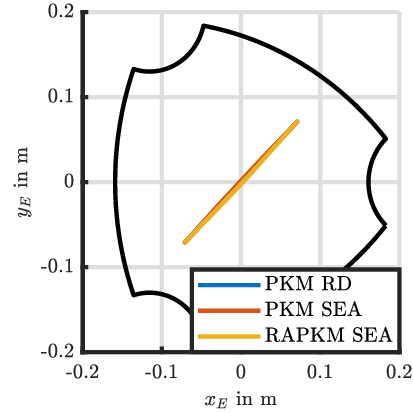}
	\captionof{figure}{Energy minimizing PKM end-effector path $\mathbf{r}_\mathrm{E} = [x_\mathrm{E},y_\mathrm{E}]^T$ of PKMs in absence of friction}
	\label{fig::PaPEELowFric}
\end{minipage}
\begin{minipage}[h]{0.6\textwidth}
	\captionof{table}{Minimal energy consumption and optimal SEA stiffnesses for a pick-and-place operation of PKMs in absence of frction. The relative values are normalized with respect to PKM RD.}
	\begin{center}
		\begin{tabular}{|c||c|c|c|}
			\hline
			&$E_{\mathrm{el}}$ in $\SI{}{\watt\second}$&$E_{\mathrm{el}}/E_{\mathrm{el,PKMRD}}$ &$c^\ast_i$ in $\SI{}{\newton \meter/\radian}$\\
			\hline
			PKM RD&$0.287$&/&/\\
			\hline
			PKM SEA&$0.285$& $99.3\,\%$&\begin{tabular}{@{}c@{}}$c^\ast_1 = 0.065$\\ $c^\ast_2 = 0.04$\end{tabular}\\
			\hline
			RAPKM SEA&$0.424$& $147.7\,\%$&\begin{tabular}{@{}c@{}}$c^\ast_1 = 0.32$\\ $c^\ast_2 = 0.32$\\$c^\ast_3 = 0.32$\end{tabular}\\
			\hline
		\end{tabular}
	\end{center}
	\label{tab::PaPEnergyLowFric}
\end{minipage}

Against all assumptions, no reduction of energy consumption can be achieved, neither for the fully actuated nor for the redundantly actuated PKM. The obvious suspect is that, during optimization, local instead of global minima were found. For that reason, a more thorough analysis in terms of spring stiffnesses was conducted. A wide set of discrete values for each SEA spring was defined and for each combination the energy optimal trajectory calculated. The optimal results are shown in Figs.  \ref{fig::PKMSEAWSLowFric} \& \ref{fig::RAPKMSEAWSLowFric}. As recognizable in Fig. \ref{fig::PKM3ArmSEA}b, the observed PKM is symmetrical (depending on configuration given by sign of $\mathbf{q}_\mathrm{P}$). Therefore, the steep increase of energy costs also appears for small values of $c_1$ but Fig. \ref{fig::PKMSEAWSLowFric} is cropped for visibility reasons. The results given in Table \ref{tab::PaPEnergyLowFric} are the global optimal solutions of Figs.  \ref{fig::PKMSEAWSLowFric} \& \ref{fig::RAPKMSEAWSLowFric}.

\begin{minipage}[t]{0.495\textwidth}
	\centering	
	\includegraphics{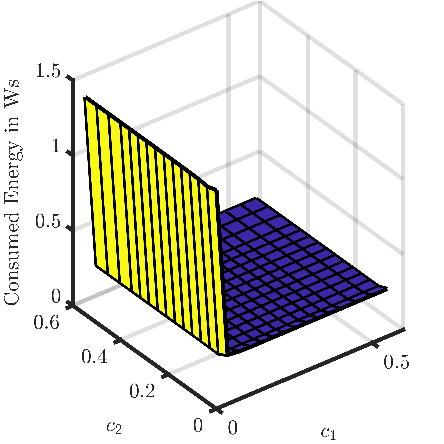}
	\captionof{figure}{Minimized energy consumption over $c_1$, $c_2$\\ in $\SI{}{\newton \meter / \radian}$ for PKM with SEAs in absence of friction}
	\label{fig::PKMSEAWSLowFric}
\end{minipage}
\begin{minipage}[t]{0.495\textwidth}
	\centering	
	\includegraphics{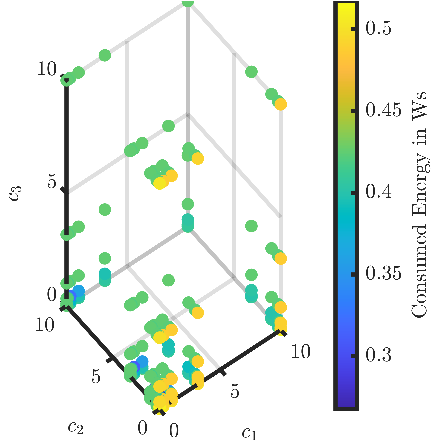}
	\captionof{figure}{Minimized energy consumption over $c_1$, $c_2$, $c_3$ in $\SI{}{\newton \meter / \radian}$ for redundantly actuated PKM with SEAs in absence of friction}
	\label{fig::RAPKMSEAWSLowFric}
	\vspace{0.35cm}
\end{minipage}

These results indicate that for the PKM depicted in Fig. \ref{fig::PKM3ArmSEA} and chosen simulation parameters no free eigenmotions can be excited. The optimal solutions converge to that of the PKM with rigid drives, because in absence of friction, eigenmotions and spring-loads, equation (\ref{equ::EOMAll}) effectively collapses towards equation (\ref{equ::EOMRD}). This hypothesis is backed up by the almost identical evolutions of the motor torques $\bm{\tau}$ in Fig. \ref{fig::PaPMLowFric}. The optimization was further conducted with base functions for $\mathbf{r}_\mathrm{E}$ based on polynomials and Fourier series but no noticeable differences were observed. In contrast to Sec. \ref{sec::SecIII}, redundant actuation acts more as a burden, adding more weight and inertia without being able to accomplish its intended purpose.

\subsection{Dominant Friction in Actuation}
A more realistic situation is that dominant friction according to (\ref{equ::EOMAll}) is present in the actuators. It is investigated in the following how the system properties and the optimization results change if friction is present. Figs. \ref{fig::PaPEEHighFric} - \ref{fig::RAPKMSEAWSHighFric} and Table \ref{tab::PaPEnergyHighFric} display similar comparisons as in the previous section. Here, in contrast to the simulation results of the previous section, the use of SEAs allows for the reduction of energy consumption with respect to rigid actuation. Around $~15\,\%$ of electrical energy can be saved throughout one cycle of the desired pick-and-place operation. As friction predominantly occurs in the actuators, performance of rigid drive actuation seems to be more affected than compliant actuation. Redundant actuation again results in higher energy consumption. The additional weight and inertia still requires more energy than it can benefit the system's properties.\vspace{0.3cm}
\begin{minipage}[h]{0.4\textwidth}
	\centering	
	\includegraphics[width=0.95\textwidth]{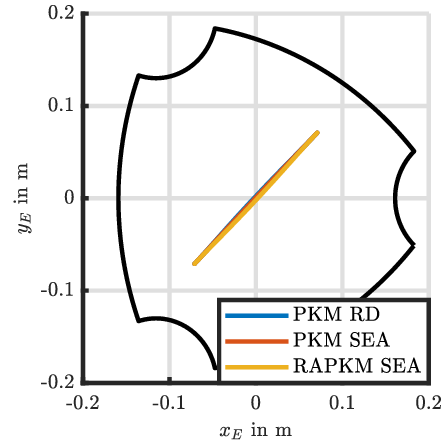}
	\captionof{figure}{Energy minimizing PKM end-effector path $\mathbf{r}_\mathrm{E} = [x_\mathrm{E},y_\mathrm{E}]^T$ of PKMs with dominant friction in the actuation}
	\label{fig::PaPEEHighFric}
\end{minipage}
\begin{minipage}[h]{0.6\textwidth}
	\captionof{table}{Minimal energy consumption and optimal SEA stiffnesses for a pick-and-place operation of PKMs with dominant friction in the actuation. The relative values are normalized with respect to PKM RD.}
	\vspace{-2ex}
	\begin{center}
		\begin{tabular}{|c||c|c|c|}
			\hline
			&$E_{\mathrm{el}}$ in $\SI{}{\watt\second}$&$E_{\mathrm{el}}/E_{\mathrm{el,PKMRD}}$ &$c^\ast_i$ in $\SI{}{\newton \meter/\radian}$\\
			\hline
			PKM RD&$0.348$&/&/\\
			\hline
			PKM SEA&$0.291$& $83.6\,\%$&\begin{tabular}{@{}c@{}}$c^\ast_1 = 0.14$\\ $c^\ast_2 = 0.18$\end{tabular}\\
			\hline
			RAPKM SEA&$0.422$& $121.3\,\%$&\begin{tabular}{@{}c@{}}$c^\ast_1 = 0.5$\\ $c^\ast_2 = 0.46$\\ $c^\ast_3 = 0.032$\end{tabular}\\
			\hline
		\end{tabular}
	\end{center}
	\label{tab::PaPEnergyHighFric}
\end{minipage}

\begin{figure}[htb]
	\centering	
	\includegraphics{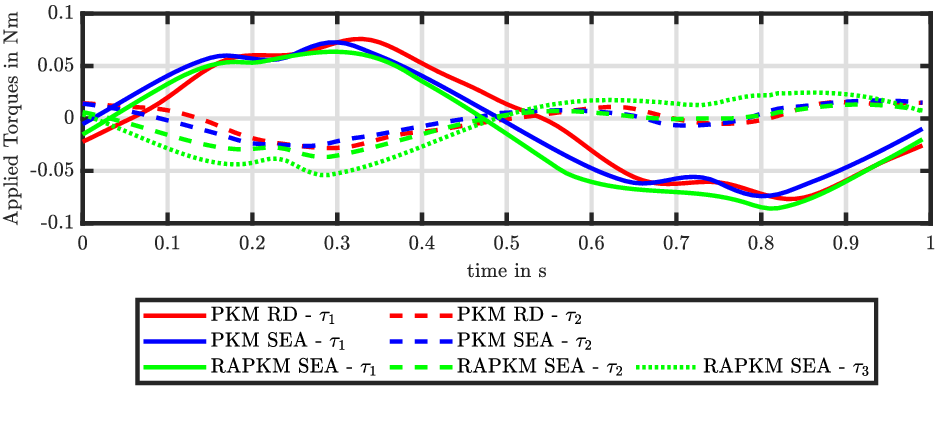}
	\caption{Energy minimized motor torques $\bm{\tau}$ of PKMs with dominant friction in the actuation}
	\label{fig::PaPMHighFric}
\end{figure}
\begin{figure}[htb]
	\centering	
	\includegraphics{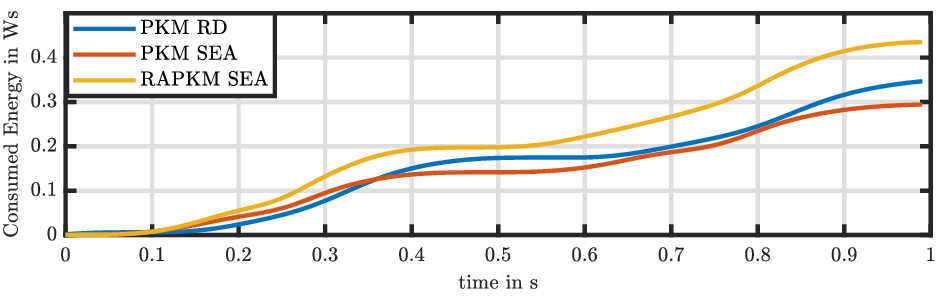}
	\caption{Consumed energy $E_\mathrm{el}$ over one cycle of PKMs with dominant friction in the actuation}
	\label{fig::PaPEnergyHighFric}
\end{figure}

\begin{minipage}[t]{0.495\textwidth}
	\centering	
	\includegraphics{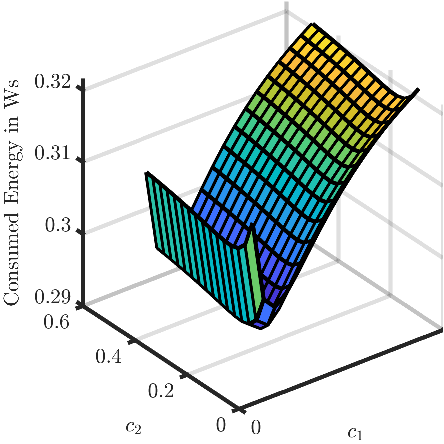}
	\captionof{figure}{Minimized energy consumption over $c_1$, $c_2$\\ in $\SI{}{\newton \meter/\radian}$ for PKM with SEAs and dominant friction\\ in the actuation}
	\label{fig::PKMSEAWSHighFric}
\end{minipage}
\begin{minipage}[t]{0.495\textwidth}
	\centering	
	\includegraphics{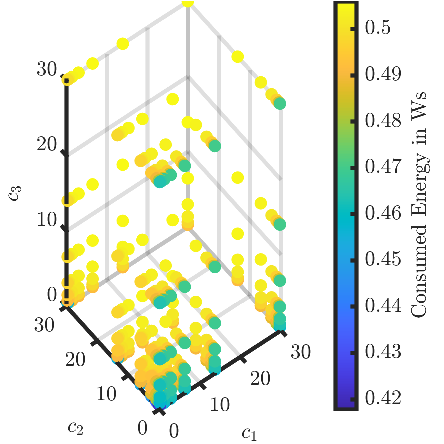}
	\captionof{figure}{Minimized energy consumption over $c_1$, $c_2$, $c_3$ in $\SI{}{\newton \meter/\radian}$ for redundantly actuated PKM with SEAs and dominant friction in the actuation}
	\label{fig::RAPKMSEAWSHighFric}
\end{minipage}

\subsection{High Actuator Inertia Compared to Load Inertia}
\begin{figure}[htb]
	\centering	
	\includegraphics{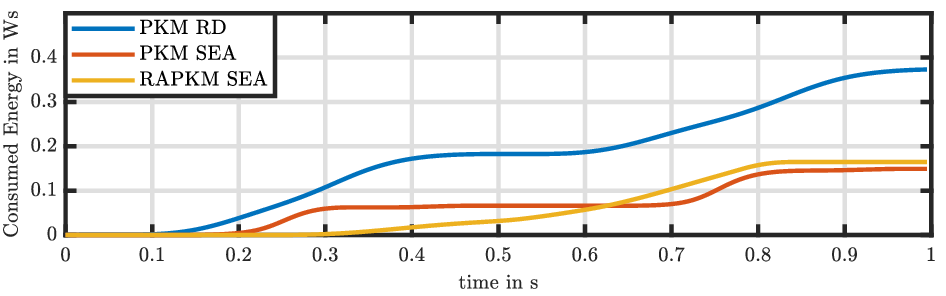}
	\caption{Consumed energy $E_\mathrm{el}$ over one cycle of PKMs with comparable actuator and load inertia}
	\label{fig::PaPEnergyHighInertia}
\end{figure}

Comparing the simulation parameters of the previous two sections with those of Sec. \ref{sec::SecIII}, a significant difference in the ratio between actuator and load inertia becomes obvious. Exciting natural oscillations, where the motor inertia is by a factor of 100 smaller than the load inertia, is intuitively difficult. In the following simulations, comparable inertias for the PKM and actuators are assumed and friction is omitted. The results are shown in Figs. \ref{fig::PaPEEHighInertia}-\ref{fig::PaPEnergyHighInertia} and Table \ref{tab::PaPEnergyHighInertia}. Here, remarkable conclusions similar to the motivation example can be drawn. Depicted in Fig. \ref{fig::PaPEEHighInertia}, for both the fully actuated PKM and redundantly actuated PKM a reduction of the consumed energy of around $60\%$ can be achieved. Furthermore, compared to a rigidly driven PKM with smaller sized motors in Sec. \ref{sec::SecV_I} even with the use of a larger sized SEAs the energy consumption can be lowered to around $55\%$. Comparable results apply to studies where friction occuring predominantly in the actuator is considered.

\begin{minipage}[h]{0.39\textwidth}
	\centering	
	\includegraphics[width=0.95\textwidth]{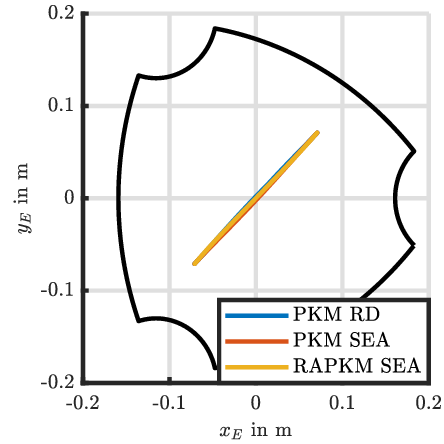}
	\captionof{figure}{Energy minimizing end-effector path $\mathbf{r}_\mathrm{E} = [x_\mathrm{E},y_\mathrm{E}]^T$ of PKMs with comparable actuator and load inertia}
	\label{fig::PaPEEHighInertia}
\end{minipage}
\vspace{0.35cm}
\begin{minipage}[h]{0.6\textwidth}
	\captionof{table}{Minimal energy consumption and optimal SEA stiffnesses for a pick-and-place operation of PKMs with comparable actuator and load inertia. The relative values are normalized with respect to PKM RD.}
	\vspace{-2ex}
	\begin{center}
		\begin{tabular}{|c||c|c|c|}
			\hline
			&$E_{\mathrm{el}}$ in $\SI{}{\watt\second}$&$E_{\mathrm{el}}/E_{\mathrm{el,PKMRD}}$ &$c^\ast_i$ in $\SI{}{\newton \meter/\radian}$\\
			\hline
			PKM RD&$0.373$&/&/\\
			\hline
			PKM SEA&$0.149$& $40\,\%$&\begin{tabular}{@{}c@{}}$c^\ast_1 = 0.03$\\ $c^\ast_2 = 0.05$\end{tabular}\\
			\hline
			RAPKM SEA&$0.165$& $44.1\,\%$&\begin{tabular}{@{}c@{}}$c^\ast_1 = 0.14$\\ $c^\ast_2 = 0.018$\\ $c^\ast_3 = 0.017$\end{tabular}\\
			\hline
		\end{tabular}
	\end{center}
	\label{tab::PaPEnergyHighInertia}
\end{minipage}

\begin{figure}[htb]
	\centering	
	\includegraphics{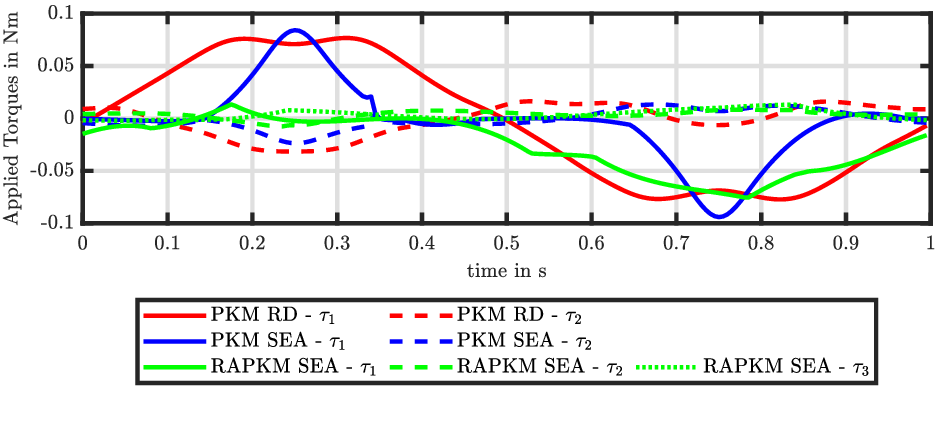}
	\caption{Energy minimized motor torques $\bm{\tau}$ of PKMs with comparable actuator and load inertia}
	\label{fig::PaPMHighInertia}
\end{figure}
\vspace{-1cm}

\section{Conclusion}\label{SecV}
In this paper, the use of SEAs and subsequent possibilities of energy savings during pick-and-place operations were explored. An offline optimal control problem with the goal to minimize energy consumption was proposed and validated in simulations. Two major conclusion could be drawn from these findings. On the one hand, the results show that the consumed energy during a cycle of a pick-and-place task can be significantly reduced if SEAs are employed for actuation. Eigenmotions can be excited and exploited, which can be further improved by redundant actuation. However, this procedure is not guaranteed to work for all robot applications. Whether it is advantageous depends on the robot kinematics and moment of inertia ratio between actuators and loads. On the other hand, another property of SEAs can be used to reduce energy consumption even in the case that no eigenmotions can be excited. The dominant amount of friction commonly appears withing an actuator. Here, adapted trajectories can be found that also penalize actuator movement and consequently lead to further energy savings. 
A way to further push this research, is the use of springs with nonlinear stiffness characteristics or even variable stiffness springs. This might especially help in robot systems with highly state-dependent and nonlinear kinematics. Additional interest lies in the analysis of structually compliant PKMs and the associated exploration of energy saving possibilities.

\section*{Acknowledgement}
This work was supported by the "LCM K2 Center for Symbiotic Mechatronics" within the framework of the Austrian COMET-K2 program.

\bibliographystyle{asmeconf}  
\bibliography{JCND_Kordik}

\end{document}